\documentclass[conference,commsoc]{IEEEtran}

\ifCLASSOPTIONcompsoc
  \usepackage[nocompress]{cite}
\else
  \usepackage{cite}
\fi

\usepackage[linesnumbered,ruled,vlined]{algorithm2e}
\usepackage{adjustbox}
\usepackage{booktabs}

\usepackage{tikz}
\usetikzlibrary{quantikz2}
\ifCLASSINFOpdf
\else
\fi

\usepackage{type1cm} 
\usepackage{lettrine}
\usepackage{graphicx} 
\usepackage{xcolor} 
\usepackage{url}
\usepackage{caption} 
\usepackage[normalem]{ulem}

\begin{document}
%
\title{Enhancing Network Anomaly Detection with Quantum GANs and Successive Data Injection for Multivariate Time Series}

\author{\IEEEauthorblockN{Wajdi Hammami}
\IEEEauthorblockA{Department of Computer Engineering\\
Polytechnique Montreal\\
Montreal, Quebec H3T 1J4\\
Email: wajdi.hammami@polymtl.ca}
\and
\IEEEauthorblockN{Soumaya Cherkaoui}
\IEEEauthorblockA{Department of Compute Engineering\\
Polytechnique Montreal\\
Montreal, Quebec H3T 1J4\\
Email: soumaya.cherkaoui@polymtl.ca}
\and
\IEEEauthorblockN{Shengrui Wang}
\IEEEauthorblockA{Department of Computer Science\\ Université de Sherbrooke\\
Sherbrooke, Quebec J1K 2R1\\
Email: Shengrui.Wang@usherbrooke.ca}
}

\maketitle

\begin{abstract}
Quantum computing may offer new approaches for advancing machine learning, including in complex tasks such as anomaly detection in network traffic. In this paper, we introduce a quantum generative adversarial network (QGAN) architecture for multivariate time-series anomaly detection that leverages variational quantum circuits (VQCs) in combination with a time-window shifting technique, data re-uploading, and successive data injection (SuDaI). The method encodes multivariate time series data as rotation angles. By integrating both data re-uploading and SuDaI, the approach maps classical data into quantum states efficiently, helping to address hardware limitations such as the restricted number of available qubits. In addition, the approach employs an anomaly scoring technique that utilizes both the generator and the discriminator output to enhance the accuracy of anomaly detection. The QGAN was trained using the parameter shift rule and benchmarked  against a classical GAN. Experimental results indicate that the quantum model achieves a accuracy high along with high recall and F1-scores in anomaly detection, and attains a lower MSE compared to the classical model. Notably, the QGAN accomplishes this performance with only 80 parameters, demonstrating competitive results with a compact architecture. Tests using a noisy simulator suggest that the approach remains effective under realistic noise-prone conditions.

Index Terms — Quantum computing, Generative adversarial networks (GANs), Anomaly detection, Variational quantum circuits (VQCs), Network security.
\end{abstract}

\IEEEpeerreviewmaketitle
\section{Introduction}

\lettrine[lines=2,slope=4pt,nindent=4pt]{A}{nomaly} detection plays a vital role across numerous domains by enabling continuous monitoring of system behavior and providing early warnings of potential issues, allowing for proactive intervention to address detected irregularities. This capability is particularly crucial in areas such as cybersecurity, where growing system complexity and surface attack has led to an increased risk of cybersecurity vulnerabilities in recent years. 6G networks will face an even broader attack surface—from massive IoT deployments and AI-orchestrated services to potential quantum-powered adversaries—demanding quantum-resistant encryption, zero-trust architectures, secure network slicing, and AI-driven threat analytics to safeguard next-generation connectivity \cite{filali2022dynamic, abouaomar2022federated, cherkaoui2021landscape}.

In this paper, we address the challenge of network anomaly detection by leveraging multi-variate time series to effectively capture the temporal dynamics present in network data. To develop an efficient anomaly detection system, the approach presented in this paper draws on two important fields that have shown considerable promise. The first is machine learning, which excels at identifying patterns and extracting insights.  This has made it a key tool of anomaly detection, with models such as Generative Adversarial Networks (GANs) showing particular effectiveness for this task due to their powerful generative capabilities \cite{luer2022anomaly, wang2021machine, IEEEhowto:dawoud, IEEEhowto:hu, AlGhuwairi2023, BlazquezGarcia2021}.

The second field is quantum computing, which has the potential to tackle computational problems that are intractable for classical computers \cite{cherkaoui2023quantum, mlika2023user, ngo2023survey, keyela2025open, Kalfon2023, PerezSalinas2019, Cerezo2021, aaraba2024quack, vieloszynski2024latentqgan, schuld2014introduction}. Quantum computing uses two fundamental principles: superposition and entanglement. Superposition allows qubits—the basic units of quantum information—to exist in multiple states simultaneously, increasing computational parallelism. Entanglement enables qubits to share correlations that can be leveraged to perform certain computations more efficiently than classical systems.

Integrating quantum computing with machine learning can potentially address challenges related to high computational overhead. Notably, some quantum machine learning approaches, especially those that employ generative models, have in some cases demonstrated advantages over classical methods, such as improved stability of the loss function during training \cite{Mirza2014, Huang2021, Rath2024, Ranga2024, Schuld2019}.

Much work has been done on network anomaly detection using machine learning \cite{hamhoum2024fortifying, moudoud2023strengthening, tashman2024, tashman2024federated, tashman2023securing, hamhoum2024mistralbsm, baseri2024, rezgui2011detecting, moudoud2020prediction, moudoud2023empowering, moudoud2023enhancing}. For instance, the authors in \cite{IEEEhowto:dawoud} examine the opportunities and challenges associated with deep learning techniques for anomaly detection, while recent studies have explored the use of self-attention mechanisms \cite{IEEEhowto:hu, hamhoum2024mistralbsm, hamhoum2024fortifying}. 

Quantum Machine Learning (QML) has emerged as a promising field, aiming to harness quantum computing for machine learning tasks \cite{schuld2014introduction}. However, despite its potential, QML faces several challenges, including hardware limitations, noise sensitivity, and the difficulty of encoding high-dimensional classical data into quantum representations  \cite{ngo2023survey}. In the context of network anomaly detection, the use of QML remains relatively unexplored \cite{ Kalfon2023, vieloszynski2024latentqgan, aaraba2024quack}, particularly with quantum generative adversarial networks (QGANs), as only a few works examining their potential \cite{Kalfon2023}. This highlights the gap in current research and underscores the novelty of this approach which applies a QGAN  to detect anomalies in multivariate network time-series data.

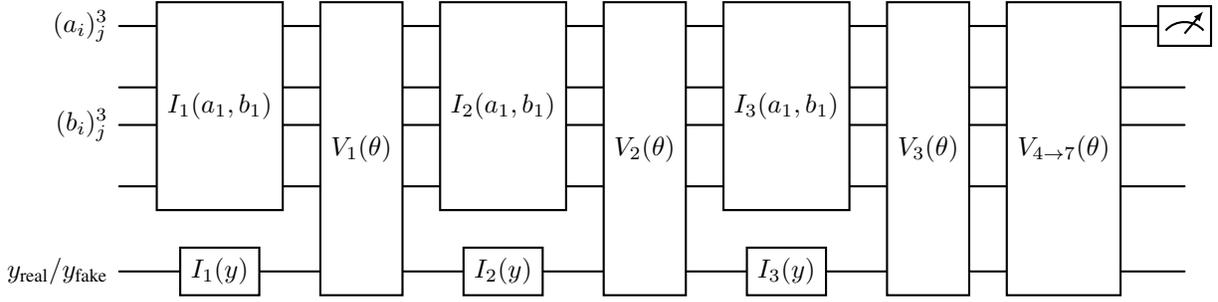
\begin{figure*}[!tb]
\centering
\begin{quantikz}
\lstick{$(a_i)_{j}^{3}$} & \gate[4]{I_1(a_1, b_1)} & \gate[5]{V_1(\theta)} & \gate[4]{I_2(a_1, b_1)} & \gate[5]{V_2(\theta)} & \gate[4]{I_3(a_1, b_1)} & \gate[5]{V_3(\theta)} & \gate[5]{V_{4\to7}(\theta)} & \meter{} \\
\lstick{} & & & & & & & & \\
\lstick{$(b_i)_{j}^{3}$} & & & & & & & & \\
\lstick{} & & & & & & & & \\
\lstick{$y_{\text{real}} / y_{\text{fake}}$} & \gate{I_1(y)} & & \gate{I_2(y)} & & \gate{I_3(y)} & & &
\end{quantikz}
\caption{Discriminator circuit}
\end{figure*}

In this work, we build upon existing approaches by designing both the generator and discriminator as quantum models. By leveraging successive data injection \cite{Kalfon2023} jointly with data re-uploading techniques \cite{PerezSalinas2019}, we extend the framework to multivariate time series. The model is trained on a real-world network traffic dataset \cite{CSE-CIC-IDS2018, Leevy2020}, and to mitigate noise, we optimize the training process by separately training the generator and discriminator, thereby reducing the number of qubits required in a single circuit. By the end of the study, we demonstrate that the model performs on par with state-of-the-art models used on the same dataset and attack types. Morover, the model achieves this
performance with only 80 parameters, demonstrating competitive results with a compact architecture.  

\section{Methodology}
\subsection{Data Preparation and Preliminary Analysis}
In this study, we use the CIC-CIDDS-2018 dataset \cite{CSE-CIC-IDS2018}, a collaborative project between the Communications Security Establishment (CSE) and the Canadian Institute for Cybersecurity (CIC). Designed to simulate a realistic network environment, the dataset captures over 80 features, providing detailed metrics such as flow duration, packet counts, and byte volumes. It includes six types of attacks: brute-force, DoS, web attacks, infiltration, botnet, and DDoS+PortScan. The dataset is publicly available through the CIC and AWS Open Data Registry. This accessibility ensures that the experimental results can be reproduced and compared with state-of-the-art methods\cite{Leevy2020}.

\noindent
\
Before conducting advanced modeling, we performed a preliminary analysis to select the most informative features, a critical necessity as qubit resources are scarce on NISQ platforms. Even with successive data injection (detailed later), quantum circuits remain highly sensitive to noise, constraining the volume of tolerable data.  

To meet the objective, we applied the Granger causality test a method for determining whether past values of one time series can predict future values of another \cite{AlGhuwairi2023}. We used it to identify significant dependencies between various network metrics and their labels, with the goal of selecting a minimal set of features that best captures anomalous behavior. The analysis is confined to the DDoS attack day (Thursday, February 20, 2018), using one-second aggregated traffic intervals.

\noindent
Taking into account both domain expertise and the constraints of today’s quantum devices—where each added input dimension deepens and noisifies the circuit—we narrowed the model to two core features:

\begin{itemize}
\item \textbf{Forward Inter-Arrival Time}: During a port scan aimed at evading detection, an increase in inter-arrival time is often observed. Attackers may deliberately slow down packet transmission to avoid triggering detection systems based on high-frequency traffic.

\item \textbf{Forward Packet Length}: This measures the size of transmitted packets. DDoS attacks frequently involve large payloads, which increase the average packet size.
\end{itemize}

\subsection{Model Architecture}

Given the constraints of quantum circuits —particularly the limited number of qubits and the sensitivity to circuit depth— we adopted a sliding window approach to manage temporal data efficiently \cite{BlazquezGarcia2021}. Instead of processing long sequences, we analyzed fixed-size windows of $\tau$ =3 at each iteration. For the multivariate time series, this involved extracting $\tau$ consecutive values from each variable and encoding them as rotation angles within the quantum circuit. This method balances the need to capture temporal dependencies with the practical limitations of current quantum hardware.
\noindent
In the model, the time window serves as the conditional input for the GAN, both at the generator and the discriminator. Rather than using pure noise as input, we control the generator’s output by providing contextual data. Similarly, the discriminator receives the same input as the generator, along with the next data point to determine whether it is real or fake. By leveraging a conditional GAN \cite{Mirza2014}, we reduce the temporal dimensionality of the data while ensuring controlled generation of the next data point.

The proposed QGAN employs Variational Quantum Circuits (VQCs), which are fundamental to hybrid quantum-classical algorithms, especially in QML. VQCs consist of parameterized quantum gates optimized through classical methods, enabling learning on  quantum devices  VQCs are composed of three parts:

a) Encoding layer: This is the part of the circuit where classical input data is encoded. There are several techniques for input encoding, including basis encoding, amplitude encoding, and angle (rotation) encoding \cite{Rath2024}\cite{Ranga2024}. In the model, we use angle encoding, i.e. we take each input value (i.e., each timestamp in the time window), transform it into a rotation angle using the $arccos$ function, and then encode it into the circuit via a rotation gate about the Y-axis of the Bloch sphere. We chose this technique because it pairs well with SuDaI \cite{Kalfon2023}. Since SuDaI trades off the number of qubits used with the circuit depth, selecting an encoding technique that uses a minimal number of gates is crucial for limiting the amount of noise.

b) Variational Layer: This is the core of the variational circuit \cite{Cerezo2021}. It consists of parameterized quantum gates followed by an entangling layer. The parameterized gates transform the encoded inputs into expressive quantum states, while the entanglement layer interconnects the qubits, effectively integrating different pieces of information from the input. In the implementation, we use two rotation gates —one for the X-axis and another for the Z-axis— which allow us to explore the entire surface of the Bloch sphere, as two rotations are sufficient. For the entanglement layer, we employ CX gates to entangle every pair of adjacent qubits, where the last qubit is entangled with the first.

a) Measurement Layer: This is the final stage of the circuit, where quantum states collapse into classical values. We use the expectation value of the Pauli Z operator as the output of the model.

\noindent
Usually, in VQCs, the variational layer (Ansatz) is repeated multiple times within a circuit to increase the model's expressive power. In this work, we employ a technique called SuDaI\cite{Kalfon2023}, where the authors not only repeat the Ansatz but also the input layer. In each new input layer, the successive timestamps are injected allowing the model to process temporal dependencies more effectively. In \cite{Kalfon2023}, the successive data injection technique is applied to a univariate time series. In the work, we extend this approach by adapting it to a multivariate time series. 

In this work, since we use a QGAN, we use two VQC, one for the generator and one for the discriminator. The generator circuit uses four qubits, while the discriminator uses six qubits. For the former, the first two qubits represent the first feature, $a_{1}$,  of the time series. The last two qubits encode the second feature, $b_{1}$, following the same pattern. A series of variational circuits follows after encoding. We alternate between an input layer and a variational layer three times. In each iteration $i$, we encode $a_{i}$, the next timestamp, (respectively $b_{i}$ for 3rd and 4th qubit). 
For the discriminator, we follow the same approach but introduce a fifth and sixth qubits, where we encode the predicted/real timestamp. Instead of using SuDaI, we employ data re-uploading \cite{PerezSalinas2019} for these additional qubits. Figure 1 illustrates the structure of the discriminator.

\subsection{Training settings and process}

The time series consists of a total of 32,401 entries. To set up the training process, we first divide the time series into two parts: one for training and  one for testing. The training-to-testing ratio is 0.5.  For training phase, we use data from the second half of the day which contain no attacks. The generator and discriminator are trained exclusively on benign data to enable them to detect anomalies.

\begin{figure*}[!tb]
    \centering
    \includegraphics[width=0.8\textwidth]{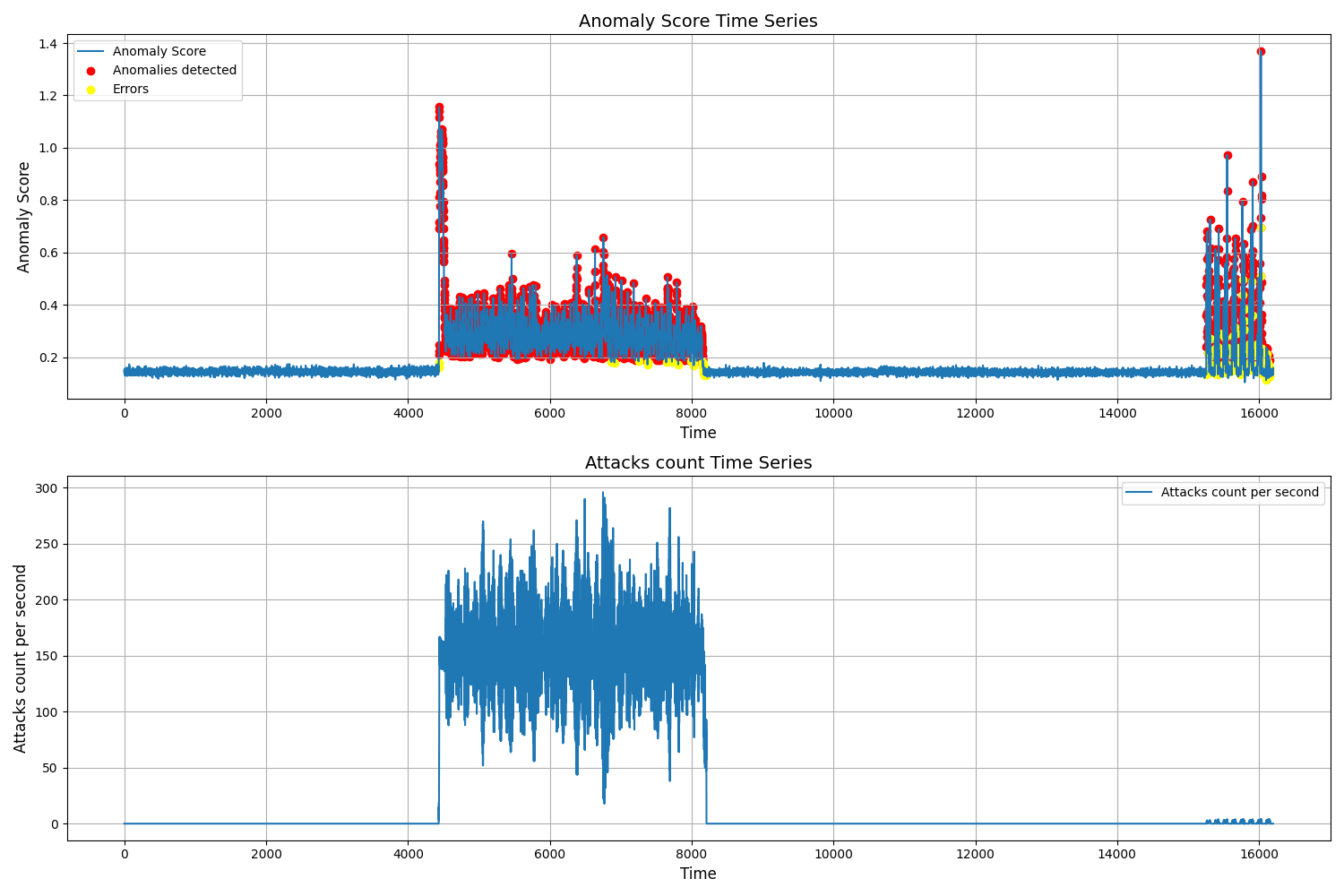} 
    \caption{Showcasing the similar patterns between the time series of attacks and the time series of anomaly scores.}
    \label{fig:feature_selection}
\end{figure*}

Prior to training, we begin by scaling the input. Since we use angle encoding,   we first scale the data to the interval $[-1, 1]$ and then apply the arccos function to obtain the corresponding angles. We follow the standard procedure used for training GANs to train the proposed QGAN. We iterate through the time series, and at each step, we take a time window of size 3, $(x_{t-2}, x_{t-1}, x_{t})$, with $x_i = [a_i, b_i]$ and generate a prediction for $x_{t+1}$. Next, we run the discriminator separately on both the real value of and the predicted value. We then compute the loss using the Mean Squared Error (MSE). The objective is for the discriminator to assign a score as close as possible to 1 for real data and 0 for fake data.  Finally, we optimize the discriminator's parameters using the parameter-shift rule\cite{Schuld2019}. 

For the generator, we generate a prediction of $x_{t+1}$ and pass it to the discriminator for scoring. We then compute the loss using the MSE, aiming for the discriminator's score to be as close as possible to 1. 

Regarding the hyperparameters, we set the learning rate to $\alpha = 0.001$.  Both the generator and discriminator were equally trained. The generator was configured with 30 parameters, while the discriminator had 50 parameters.

\subsection{Anomaly Detection}

\begin{algorithm}
\caption{Anomaly Detection Using QGAN}
\label{alg:anomaly_detection}
\SetKwInOut{Input}{Input}
\SetKwInOut{Output}{Output}

\Input{Trained Generator $G$, Trained Discriminator $D$, Time series window $X_t = \{x_{t-2}, x_{t-1}, x_t\}$, Actual target $x_{t+1}$}
\Output{Anomaly Score $S_t$}

$\hat{x}_{t+1} \leftarrow G(X_t)$

$MSE_t \leftarrow \frac{1}{n} \sum (x_{t+1} - \hat{x}_{t+1})^2$

$D_t \leftarrow D(X_t, x_{t+1})$

\
$w_G \leftarrow \frac{L_D}{L_G + L_D}$, $w_D \leftarrow \frac{L_G}{L_G + L_D}$ 
$S_t \leftarrow w_G \cdot MSE_t + w_D \cdot (1 - D_t)$

Compute anomaly scores for benign traffic $S_{benign}$

Set threshold $T$ as 99.99th percentile of $S_{benign}$

\eIf{$S_t > T$}{
    Flag $x_{t+1}$ as anomaly
}{
    Mark $x_{t+1}$ as normal
}

\end{algorithm}


We leverage both the generator and discriminator of the trained model for anomaly detection, which is the main objective. The core idea behind our approach is to use the generator for predictive modeling and the discriminator for distinguishing between normal and anomalous behavior. By combining their outputs, we create a robust anomaly detection mechanism.

To achieve this, we compute an anomaly score based on the model’s outputs.
Specifically, the generator takes the time window preceding the target data point to generate a prediction. We then compute MSE between the predicted and actual values.

For the discriminator, we provide it with the time window and the actual target value to determine whether it aligns with the distribution of benign traffic.
Finally, we combine the MSE and discriminator output through a weighted summation. The weights are determined based on the generator and discriminator losses from the last training iteration, assigning higher confidence to the component with the lower loss.
\begin{table*}[h!]
\centering
\caption{Comparison of different detection methods of the model}
\begin{adjustbox}{max width=1\textwidth}
\begin{tabular}{lcccccc}
\toprule
\textbf{Detection Method} & \textbf{Accuracy} & \textbf{Recall} & \textbf{Precision} & \textbf{F1-Score} & \textbf{MSE} \\
\midrule
QGAN Discriminator-Only Detection & 0.83 & 0.35 & 0.98 & 0.51 & 0.0013 \\
QGAN Generator-Only Detection     & 0.26 & 1.00 & 0.26 & 0.41 & 0.0013 \\
QGAN Combined Detection           & 0.99 & 0.97 & 0.99 & 0.98 & 0.0013 \\
Classic GAN Combined Detection                & 0.96 & 0.86 & 0.99 & 0.92 & 0.0157 \\
Noisy QGAN Combined Detection               & 0.97 & 0.89 & 0.99 & 0.94 & 0.0042 \\
\bottomrule
\end{tabular}
\end{adjustbox}
\end{table*}

\begin{table}[h!]
\centering
\caption{Evaluation results of deep learning models for DDoS detection \cite{Agostinello2023}}
\begin{adjustbox}{max width=0.5\textwidth}
\begin{tabular}{lcccccc}
\toprule
\textbf{Model} & \textbf{Accuracy} & \textbf{Precision} & \textbf{Recall} & \textbf{F1-Score} & \textbf{Trainable Parameters} \\ \midrule
DNN            & 98.80\%           & 98.80\%            & 98.80\%         & 98.80\%          & 2,977                                   \\
CNN            & 99.33\%           & 99.33\%            & 98.34\%         & 99.33\%          & 8,065                                 \\
RNN            & 99.78\%           & 99.78\%            & 99.78\%         & 99.78\%          & 12,705                                  \\

\bottomrule
\end{tabular}
\end{adjustbox}
\end{table}

To determine the threshold for anomaly detection, we first calculate the anomaly scores from the benign traffic (training data). We then compute the 99.99th percentile, meaning only the top 0.01\% of benign values exceeds this threshold, making them potential outliers or anomalies. It should be noted that the threshold value of 99.99\% here is user-fixed. Automatic determination of this threshold is not addressed in this work. Algorithm 1 provides a detailed illustration of this process.

\section{Results \& Discussion}

To evaluate the proposed model, we conducted a series of tests to assess each component's contribution. We first trained a classical GAN using the same number of parameters as the QGAN, which resulted in poor performance. Competitive results were only achieved after increasing the classical GAN's parameters to 51 for the generator and 55 for the discriminator. 

To test the generative power of the QGAN, we first trained it with a synthetic multivariate time series combining a random walk with a modulated sinusoidal wave. The first variable introduces randomness, while the second adds periodic complexity through amplitude and phase modulation. The results of the QGAN are illustrated in Figure 3, while the results of the clasical GAN are illustrated in Figure 4 where both models were trained for the same number of epochs.

The QGAN achievs an MSE of 0.015, significantly outperforming the classical GAN’s MSE of 0.078.  The QGAN better captures fine details for this time series, particularly noise fluctuations, highlighting its superior generative capability compared to its classical counterpart.

\begin{figure}[!tb] 
\centering \includegraphics[width=0.5\textwidth]{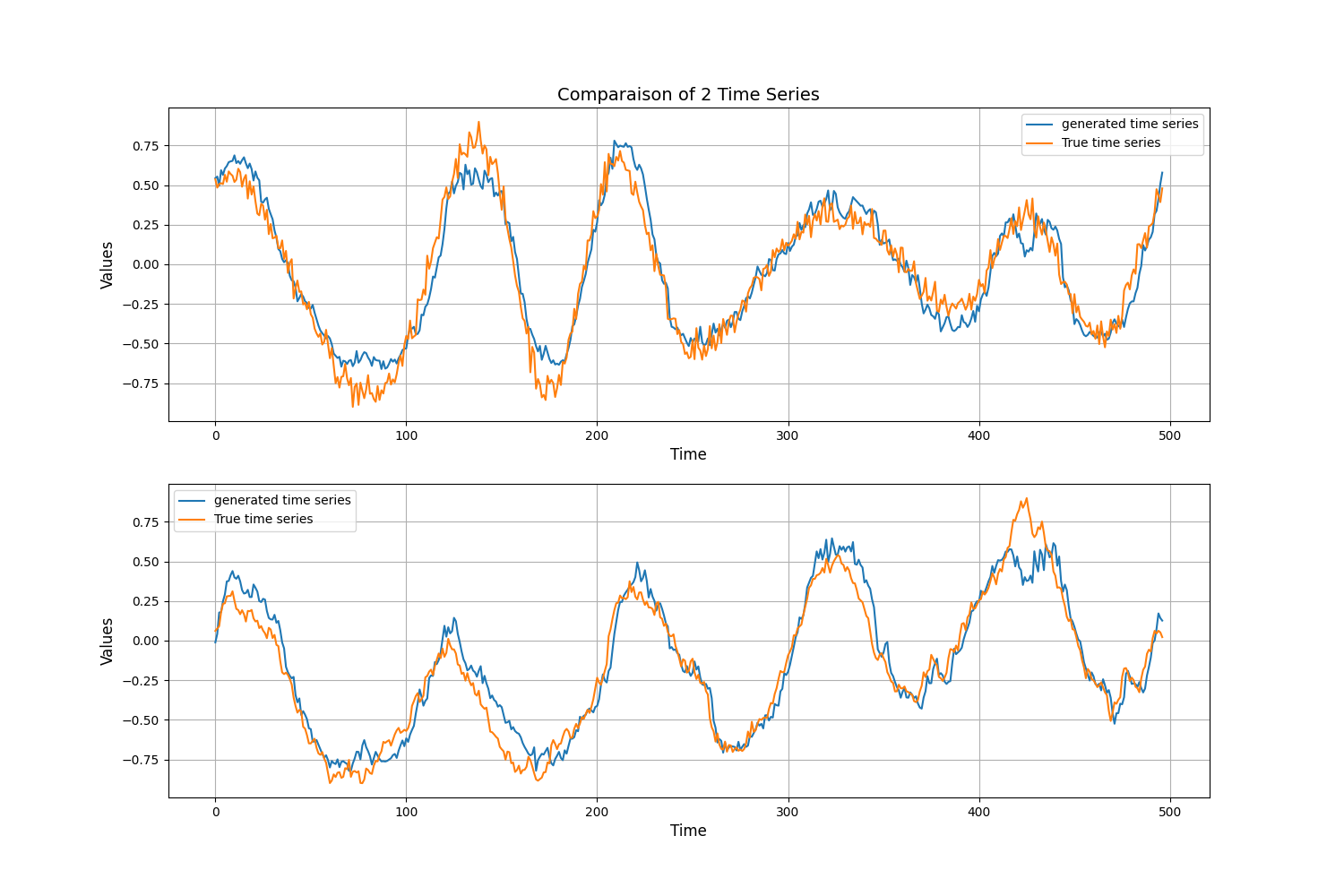} 
\caption{QGAN results for synthetic time series.} \label{fig:feature_selection} 
\end{figure}

\begin{figure}[!tb] 
\centering \includegraphics[width=0.5\textwidth]{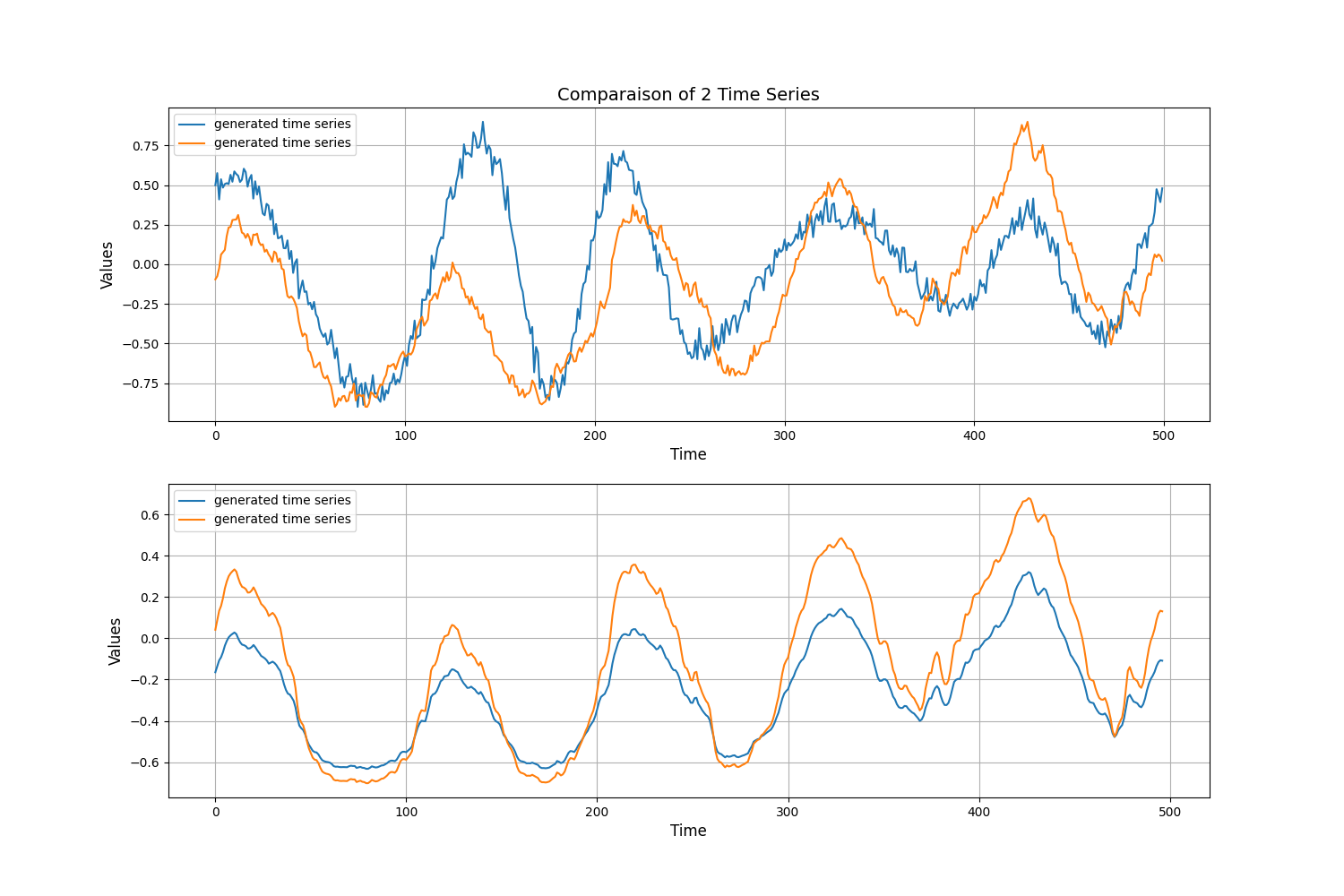} 
\caption{Classic GAN results for synthetic time series.} \label{fig:feature_selection1} 
\end{figure}

Following this, we assessed the model's ability to detect anomalies for the real dataset by evaluating three approaches: the generator alone (Generator-Only Detection), the discriminator alone (Discriminator-Only Detection), and a combined approach (Combined Detection). Moreover, we also assessed the combined approach of the QGAN  with a noisy simulator using the Fake Vigo backend in order to evaluate the model’s robustness under realistic quantum noise conditions. This setup allows us to simulate the behavior of a quantum device affected by noise, providing preliminary insights into the performance of the QGAN when deployed on actual quantum hardware proving a good foundation for future hardware deployment. The results in terms of accuracy, recall, precision, F1-score and MSE for all approaches are shown in Table I

For each test, we compute standard metrics such as accuracy, recall, and F1-score, along with MSE for the training set. The MSE measures how well the generator predicts benign time-series data.
The results indicate that the combined approach (Combined Detection), which integrates both the generator and the discriminator, achieves superior performance. The approach attains an accuracy of 0.99 and an F1-score of 0.98, along with the lowest MSE of 0.0013. In contrast, the discriminator-only approach (Discriminator-Only Detection) shows high precision (0.98) but suffers from low recall (0.35) which indicates that it misses many anomalies. The generator-only approach (Generator-Only Detection) achieves perfect recall (1), but its low precision (0.26) suggest that it produces numerous false positives. The classical GAN however, even after increasing the number of parameters, achieves competitive performance (accuracy of 0.96, recall of 0.86, precision of 0.99, and F1-score of 0.92) but exhibits a significantly higher MSE of 0.0157. Moreover, testing the quantum model on a noisy simulator (Noisy anomaly) yields strong performance (accuracy 0.97,  F1-score 0.94) with an MSE of 0.0042, which demonstrate its robustness to noise.\\
\noindent
The individual performances of Discriminator-Only Detection and Generator-Only Detection highlight the effectiveness of combining both components in the QGAN model. For instance, Discriminator-Only Detection exhibits high precision, accurately identifying true positives. On the other hand, Generator-Only Detection demonstrates high recall, effectively capturing a wide range of anomalies. By integrating both, we leverage the strengths of each —Discriminator-Only Detection's precision and Generator-Only Detection's recall— resulting in a more robust and comprehensive anomaly detection system.\\
\noindent
Table II \cite{Agostinello2023} presents the results achieved by using various classical machine learning model for anomaly detection on the same data set used in this work. While deep learning models such as CNNs and RNNs achieve slightly higher accuracy, they require thousands of parameters. In contrast, the QGAN uses only 80 parameters while still delivering competitive performance. This indicates the potential of quantum models to achieve good results with significantly fewer resources in terms of number of parameters.\\
\noindent
Finally, in Figure 2, we present the results of the anomaly scores in the test data set for the QGAN using the combined approach. Yellow highlights indicate moments where the model failed to detect attacks. However, we observe that these errors still occur within time periods where attacks have already been detected. In real-world scenarios, missing these anomalies would have a minimal impact.

\section{Conclusion \& Future Work}
In this work, we introduced a QGAN for network anomaly detection in a real-world multi-variate dataset. To address qubit limitations, we optimized qubit usage by leveraging data re-uploading, and successive data injection. To enhance feature selection, we applied the Granger causality test, identifying the most relevant network features.
Experimental results show that the QGAN achieves high accuracy, recall, and F1-scores in anomaly detection, while also yielding a lower MSE compared to the classical model. Remarkably, the QGAN delivers this performance using just 80 parameters, highlighting its competitive capability with a compact architecture.

Despite the promising results, several areas offer opportunities for future enhancement. First, the model was evaluated on a single type of attack (DDoS) and a limited set of features, which may not generalize to more complex or diverse network threats. Second, we tested the model on a noisy simulator to assess the QGAN’s resilience to quantum noise. However, testing on an actual quantum computer would provide a more definitive evaluation of the model's performance under real hardware constraints, including decoherence, gate errors, and measurement noise. Such testing is essential to fully validate the model’s practical viability and guide future improvements. Lastly, the current thresholding method for anomaly detection is manually set, which may not adapt well to dynamic environments. Addressing these aspects presents valuable directions for future research and development.

\ifCLASSOPTIONcompsoc
  \section*{Acknowledgments}
\else
  \section*{Acknowledgment}
\fi

The authors gratefully acknowledge the financial support provided by NSERC, Prompt Québec, and Thales Canada.

\end{document}